\documentclass[11pt,a4paper,dvipsnames]{article}
\usepackage[hyperref]{acl2021}
\usepackage{times}
\usepackage{latexsym}
\usepackage{amsmath}
\usepackage{amssymb}

\usepackage{hyperref}
\usepackage{multicol}

\usepackage{array}
\usepackage{caption}
\usepackage{subcaption}

\usepackage{microtype}
\usepackage{xcolor}
\usepackage{graphicx}
\usepackage{soul}
\usepackage{arydshln}
\usepackage{tikz}
\usepackage{tikz-qtree}

\aclfinalcopy 

\newcommand{\pgen}{p_{\text{gen}}}
\newcommand{\ngram}{$n$-gram}

\newcommand{\pmin}{p_{\text{min}}}
\newcommand{\hgen}{H_{\text{gen}}}
\newcommand{\hcopy}{H_{\text{copy}}}
\newcommand{\hngram}{H_{\text{ngram}}}
\newcommand{\hlstm}{H_{\text{LSTM}}}
\newcommand{\hparser}{H_{\text{parser}}}
\DeclareMathOperator{\Dedge}{D_{\text{edge}}}
\DeclareMathOperator{\Droot}{D_{\text{root}}}

\DeclareMathOperator{\CC}{CC}

\title{To Point or Not to Point: Understanding How Abstractive Summarizers Paraphrase Text}

\author{Matt Wilber \\
  Cornell University \\
  \texttt{mw979@cornell.edu} \\\And
  William Timkey \\
  Cornell University \\
  \texttt{wpt25@cornell.edu} \\\And
  Marten van Schijndel \\
  Cornell University \\
  \texttt{mv443@cornell.edu}}

\date{}

\begin{document}
\maketitle
\begin{abstract}
Abstractive neural summarization models have seen great improvements in recent years, as shown by ROUGE scores of the generated summaries. But despite these improved metrics, there is limited 
understanding of the strategies different models employ, and how those strategies relate their understanding of language.
To understand this better, we run several experiments to characterize how one popular abstractive model, the pointer-generator model of \citet{see-etal-2017-get}, uses its explicit copy/generation switch to control its level of abstraction (generation) vs extraction (copying). On an extractive-biased dataset, the model utilizes syntactic boundaries to truncate sentences that are otherwise often copied verbatim. When we modify the copy/generation switch and force the model to generate, only simple paraphrasing abilities are revealed alongside factual inaccuracies and hallucinations. On an abstractive-biased dataset, the model copies infrequently but shows similarly limited abstractive abilities. In line with previous research, these results suggest that abstractive summarization models lack the semantic understanding necessary to generate paraphrases that are both abstractive and faithful to the source document.

\end{abstract}

\section{Introduction}

Recent years have seen great improvements in ``abstractive'' summarization models -- models that not only concatenate text from the source document, but can additionally paraphrase to generate summary text. Once limited to sentence compression \cite{rush-etal-2015-neural}, abstractive models now generate multi-sentence summaries \cite{see-etal-2017-get}, even for relatively long documents \cite{cohan-etal-2018-discourse}. However, extractive models and mixed models with significant extractive components continue to show strong performance, and the extent and manner in which abstraction is used by summarization models is not well understood.

Previous work has raised concerns about whether models are able to paraphrase in ways that lead to better summaries. Abstractive models often generate summaries that are either ungrammatical or unfaithful to the source document \cite{maynez-etal-2020-faithfulness, durmus2020feqa, kryscinski-etal-2020-evaluating} and are prone to repetition in their outputs \cite{see-etal-2019-massively, holtzman2020curious}.
These issues raise questions about \emph{how} neural summarizers generate novel text. Abstractive summarization is differentiated from extractive summarization by the model's ability to paraphrase, but 
paraphrasing ability is not directly measured by popular metrics, leading to a lack of understanding of the generative process. Some previous research has aimed to alleviate these issues in evaluation: \citet{zhang2018abstractiveness} propose evaluating summaries with human evaluations of informativeness and coherence, and \citet{Ganesan2018ROUGE2U} implements a metric to reward models that paraphrase via simple synonym substitutions according to WordNet. However, synonym substitution is just one form of paraphrasing, and truly abstractive models should be capable of more complex paraphrasing strategies.

To understand how abstraction manifests in neural summarization models, we study a model that has an explicit abstraction/extraction switch, the pointer-generator model of \citet{see-etal-2017-get}. The training objective of this model causes it to choose the best summarization strategy (abstractive vs extractive) in different contexts, permitting us to determine the environments where abstractive summarization is an effective summarization strategy. First, we show how the switch varies across a full summary and is influenced by the decoder's copy and generation distributions. Next, we present a behavioral probe of the abstraction/extraction switch, to observe how the switch reacts to lexical, structural, and distributional information as it decodes a summary. Finally, we modify the switch value, forcing more frequent paraphrase generation during decoding, revealing the limits of the model's paraphrasing capabilities. Ultimately, we find across both the CNN/DailyMail and XSum datasets that the model's abstractive capabilities are limited; the model understands how to identify and combine constituents from the source text in a grammatical fashion, but lacks the semantic understanding required to produce grammatical, faithful and meaningful paraphrases.

\section{Model}

\subsection{The Pointer-Generator Model}

We study the pointer-generator model released by \citet{see-etal-2017-get}, which uses an explicit switch, $\pgen$, that blends abstractive and extractive summarization strategies. We briefly review the pointer-generator model here; for more details, see the original paper of \citet{see-etal-2017-get}.

The final output distribution for a particular word in the summary $P(w)$ is a weighted sum of the generation distribution and the copy distribution, weighted by $\pgen$ and $1 - \pgen$, respectively. This is described by Equation 9 in \citet{see-etal-2017-get}, modified for clarity here:
\begin{equation}
P(w) = \pgen P_{\text{vocab}}(w) + (1-\pgen)  P_{\text{copy}}(w)
\end{equation}

$P_{\text{vocab}}(w)$ is the generation distribution over the model's vocabulary, and $P_{\text{copy}}(w)$ is the copy distribution over the tokens in the source document. The $\pgen$ switch explicitly weights the influence of the generation and copy mechanisms on $P(w)$. For each time step $t$, $\pgen$ is a function of the context vector $h_t^*$, the decoder state $s_t$ and the decoder input $x_t$,
\begin{equation}
    \pgen = \sigma(\delta_{h_t^*}^T h_t^* + \delta_s^T s_t + \delta_x^T x_t + \beta_{\text{ptr}})
\end{equation}

where $\sigma$ is the sigmoid function and $\delta_{h_t^*}$, $\delta_s$, $\delta_x$ and $\beta_{\text{ptr}}$ are learned parameters.

\citet{see-etal-2017-get} also use a coverage mechanism aimed at reducing repetition, defining the coverage vector $c^t$ as
\begin{equation}
    c^t = \sum_{t'=0}^{t-1} P_{\text{copy}(w_t)}
\end{equation}
which is passed as another input to the attention mechanism.

\subsection{Data}

We analyze pointer-generator behavior when trained on an extractive-biased dataset, CNN/DailyMail, and on an abstractive-biased dataset, XSum. The CNN/DailyMail dataset is made up of multi-sentence summaries of news articles from CNN and Daily Mail. XSum \citep{narayan2018don} is a summarization dataset that uses the first sentence of a news article as a summary of the article. The dataset treats the remainder of the article as the source document. As a result, the summaries are both shorter and more difficult to copy from the source document, compared to the CNN/DailyMail dataset.

\subsection{Training}

Our experiments on CNN/DailyMail use the trained model released by \citet{see-etal-2017-get}, which includes the coverage mechanism described above. We decode summaries on the test set of at most 120 tokens using beam search with beam width 4, as in the original paper.
For XSum, we trained our own model on the XSum training partition, using the code released by \citet{see-etal-2017-get}.\footnote{Code and full replication details are available at \href{https://github.com/mwilbz/pointer-generator-analysis}{https://github.com/mwilbz/pointer-generator-analysis}.}

Like \citet{narayan2018don}, we do not include the coverage mechanism for the XSum model. When coverage is used for the XSum model, ROUGE scores \citep{lin-2004-rouge} slightly decrease, and the produced summaries contain more severe hallucinations. However, adding coverage does ``fix'' some degenerate summaries that produce the same sequence of tokens repeatedly -- see Appendix B for an example.

For both datasets, in addition to the output summaries, we record the value of the $\pgen$ switch for each emitted token, as well as the generation distribution and the copy distribution at each time step.

\section{Experiments}\label{section:experiments}

\begin{figure*}[h!]
    \centering
    \begin{subfigure}[b]{\textwidth}
        \centering
        \includegraphics[width=\linewidth]{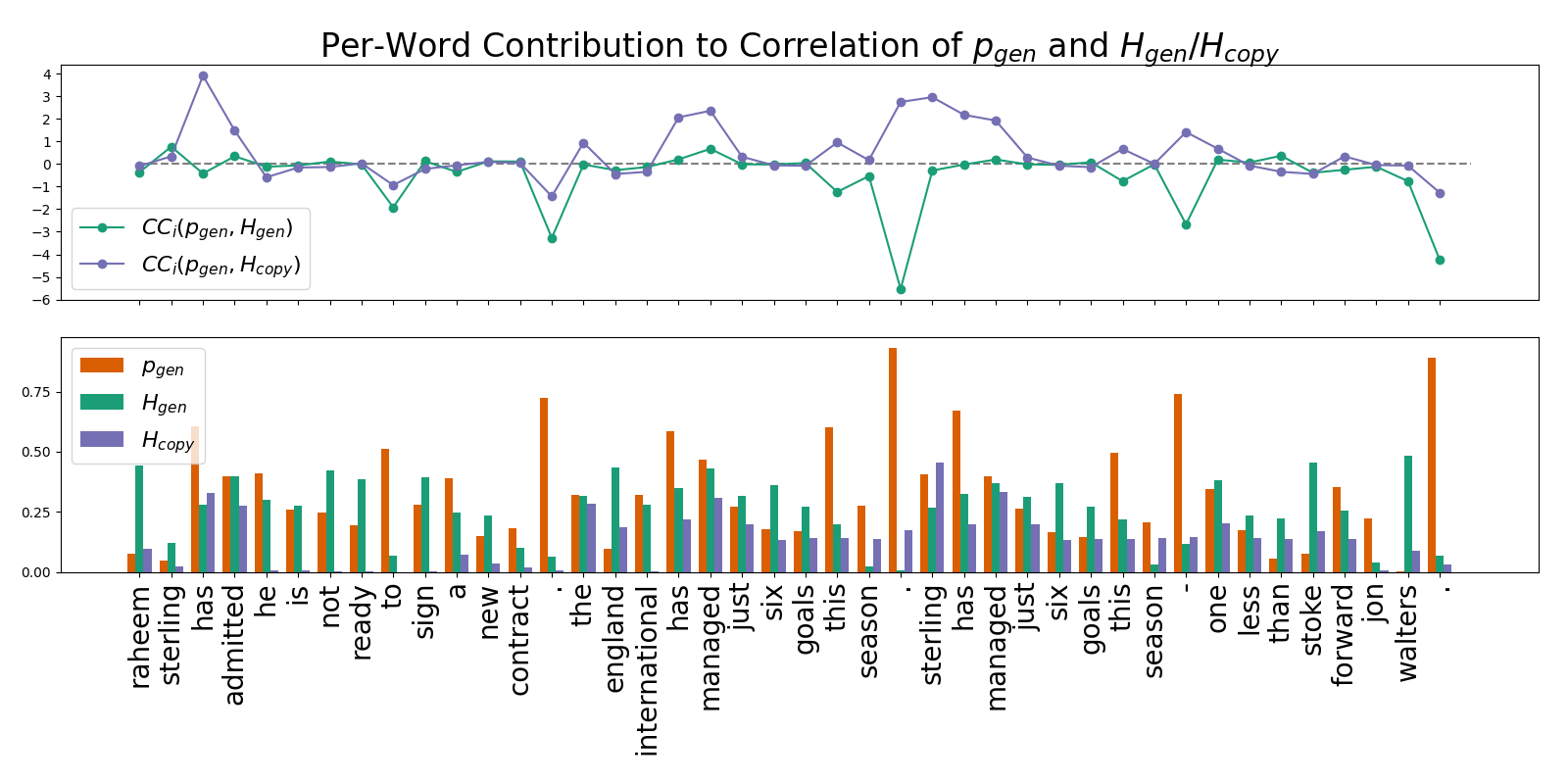}
        \caption{CNN/DailyMail}
    \end{subfigure}
    \begin{subfigure}[b]{\textwidth}
        \centering
        \includegraphics[width=\linewidth]{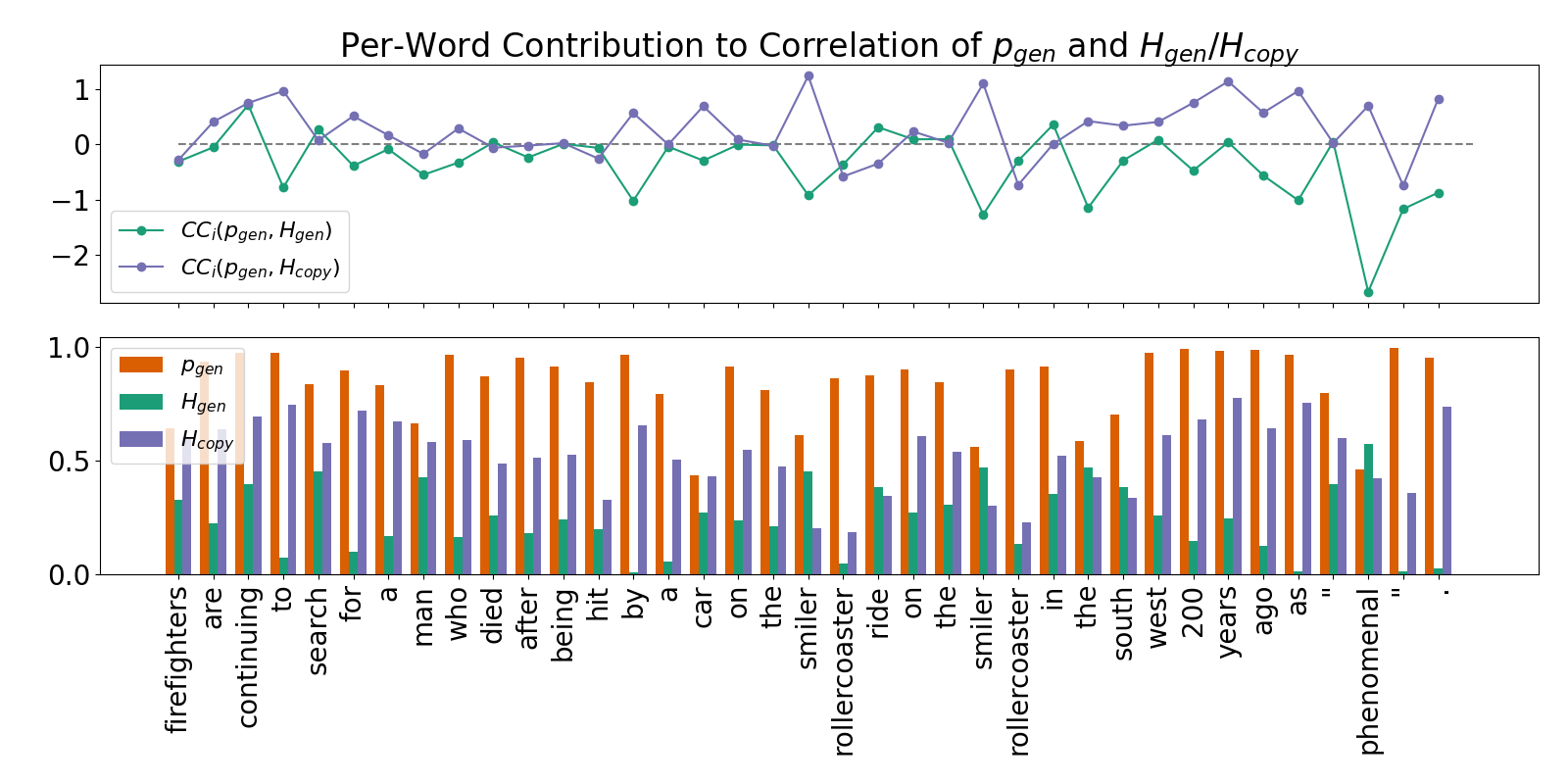}
        \caption{XSum}
    \end{subfigure}
    \caption{(Top) Correlation contributions $\CC(\pgen, \hgen)$ (green) and $\CC(\pgen, \hcopy)$ (purple) for a randomly-sampled summary. (Bottom) Bar plot of per-token $\pgen$ (orange), and entropy of the generation distribution (green) and copy distribution (purple) for the same summary.}
    \label{fig:corr_contrib_0}
\end{figure*}

In Section~\ref{section:experiment_token_level} we qualitatively analyze the evolution of the per-token $\pgen$ and uncertainty in the extractive/abstractive components over the course of randomly selected summaries. Section~\ref{section:probing_pgen} provides quantitative evidence of our observations across the full test sets, by modeling the lexical, structural, and distributional ($P_{\text{vocab}}$ and $P_{\text{copy}}$) environments that drive the variability of the ${\pgen}$ switch.

Finally, in Section~\ref{section:modifying_pgen} we manipulate $\pgen$ of the CNN/DailyMail model to generate summaries that are more abstractive than those of the base model, in order to disentangle any abstractive \emph{behavior} from abstractive \emph{capabilities}, finding that the model's abstractive capabilities are largely limited to lexical paraphrases, and that forcing the model to generate more novel text yields unfaithful summaries.

\subsection{Token-level Analysis} \label{section:experiment_token_level}

\subsubsection{Model}

The $\pgen$ switch explicitly tells us how much weight is assigned to the generation and copy distributions.
\citet{see-etal-2017-get} make qualitative claims about the environments where ${\pgen}$ is highest: “We find that ${\pgen}$ is highest at times of uncertainty such as the beginning of sentences, the join between stitched-together fragments, and when producing periods that truncate a copied sentence.” In this section, we evaluative these observations on randomly selected summaries generated with each model.

We quantify the notion of ``uncertainty'' from \citet{see-etal-2017-get} using information-theoretic entropy \cite{shannon48} of the distribution that predicts the next word ${w_i}$ of a generated summary:
\begin{equation}
    H_\theta(w_i) = \mathbb{E}_{P_\theta} \left[ -\log P_\theta(w_{i}) \right].
\end{equation}
where $P_\theta$ is the predictive distribution over the model vocabulary $V_\theta$ at a given time step. In our experiments, we use \emph{normalized} entropy, which divides the equation above by $\log_2 |V_\theta|$, to limit the domain to $[0,1]$ regardless of the vocabulary size. We calculate model-internal entropies $\hgen$ and $\hcopy$ by setting $P_\theta$ equal to $P_{\text{vocab}}$ and $P_{\text{copy}}$, respectively.

Given the entropy of the copy and generation distributions at each decoder time step, we investigate the relationship between $\pgen$, $\hgen$, and $\hcopy$ by calculating per-token \emph{correlation contributions}. Intuitively, correlation contribution measures how much an individual token contributes to either positive or negative correlation between $\pgen$ and the model entropies.

The Pearson correlation coefficient between two sequences $\mathbf{x} = [x_1, \ldots, x_n]$ and $\mathbf{y} = [y_1, \ldots, y_n]$ can be written as

\begin{equation}
    r = \frac{\sum_{i=1}^n (x_i - \bar{x}) (y_i - \bar{y})}{\sqrt{\sum_{i=1}^n (x_i-\bar{x})^2 \sum_{i=1}^n (y_i-\bar{y})^2}}
\end{equation}
We calculate the correlation contribution of the pair $(x_i, y_i)$ at index $i$ to be
\begin{equation}
    \CC_i = \frac{n (x_i - \bar{x}) (y_i - \bar{y})}{\sqrt{\sum_{i=1}^n (x_i-\bar{x})^2 \sum_{i=1}^n (y_i-\bar{y})^2}}
\end{equation}
Note that the correlation between $\mathbf{x}$ and $\mathbf{y}$ is equal to the average of $\CC_1, \CC_2, \ldots, \CC_n$, but unlike $r$, the correlation coefficient, each component $\CC_i$ is not bounded by $[-1, 1]$.

\subsubsection{Results}

Across the test splits, the Pearson correlation betwen $\pgen$ and $\hgen$ is $-0.47$ for CNN/DailyMail and $-0.55$ for XSum. The correlation between $\pgen$ and $\hcopy$ is $0.12$ for CNN/DailyMail and $0.54$ for XSum. This suggests that the higher-certainty (lower $H$) distribution is weighted more heavily when combining the generation and copy distributions, since $\pgen$ is high when $\hgen$ is low, and low when $\hcopy$ is low.

Visualizing the correlation contributions across a sentence helps us understand how individual tokens are decoded as a function of uncertainty in the abstractive and extractive components of the model. We randomly sample articles from each dataset's test split, and visualize the correlation contributions for the generated summaries in Figure \ref{fig:corr_contrib_0}. Additional examples may be found in Appendix A.

\textbf{CNN/DailyMail}: The tokens that correlate high $\pgen$ with low $\hgen$ (high certainty in the abstractive component) are frequently punctuation, and periods in particular. This punctuation appears to be used to truncate sentences at a syntactic boundary, a behavior we quantify in Section \ref{section:probing_pgen}. The correlation of high $\pgen$ and high $\hcopy$ (low certainty in the extractive component) comes from tokens including ``has'', ``managed'', ``.'', and ``sterling''; all tokens that appear multiple times in the source document. This suggests a possible role played by generation to tie break when the copy distribution has low certainty about which continuation to copy next.

\textbf{XSum}: The XSum model uses the copy mechanism very infrequently; $\pgen$ is frequently large. When $\pgen$ is small, we tend to observe uncertainty in the generative component and certainty in the copy component, according to entropy measures. In Figure \ref{fig:corr_contrib_0}, we see this happens when the proper noun ``smiler'', a rollercoaster name, is generated. It also happens at the beginning of a quotation, indicating that the model has learned that quotations should be copied from the source document, rather than generated.

Overall, we see a strong contrast in $\pgen$ values between the two models. On the extractive-biased CNN/DailyMail dataset, the model learns to copy frequently, generating where necessary to truncate sentences. On the generative-biased XSum dataset, the model acts nearly like a simple seq2seq model, only infrequently using the copy mechanism for the sake of proper nouns and quotations.\footnote{This can also be seen in the contrasting gaps between the seq2seq and pointer-generator ROGUE scores reported by \citet{see-etal-2017-get} and \citet{narayan2018don}. The former sees a 9-point gap in ROUGE-1, while the latter reports a 1-point gap.}

\subsection{Probing $\pgen$} \label{section:probing_pgen}

In the previous section, we made qualitative observations about the relationship between $\pgen$ and model entropies, as well as the linguistic environments where $\pgen$ is highest. In this section, we quantify these relationships by predicting $\pgen$ with a linear model of lexical, syntactic and distributional factors. 

\subsubsection{Model Features}
In this section, we describe the four feature sets we use to model ${\pgen}$. These include model-internal entropy measures from the \citet{see-etal-2017-get} summarizer, model-external entropy measures derived from pretrained language models, structural features derived from syntactic parses of summaries, and part-of-speech tags.

\textbf{Summarization model entropies:} We use $\hgen$ and $\hcopy$ as features, hypothesizing, like \citet{see-etal-2017-get}, that the uncertainty in the copy and generation distributions will have a significant effect on $\pgen$.

\textbf{Language model entropies:} We also use entropy from three types of language models with varying degrees of lexical and structural expressiveness: a trigram model,\footnote{A Kneser-Ney trigram model trained on 5.4m tokens of the articles from the training partition of the summarization dataset.} a top-down incremental constituency parser \cite{roark-2001-probabilistic,roark-etal-2009-deriving}, and a unidirectional recurrent neural language model \cite{van-schijndel-etal-2019-quantity}. These models allow us to directly measure how much ${\pgen}$ may be influenced by lexical, syntactic, and distributional uncertainty in the generated summary independent of the summarization objective. 

\textbf{Structural Features:} The summarization model may also condition its decision to copy or generate on the current syntactic environment. While pointer-generator models do not explicitly model syntax, they may exhibit some implicit syntactic knowledge, such as the ability to identify and copy whole constituents. As mentioned above, \citet{see-etal-2017-get} claim that $\pgen$ is high at the “the join between stitched-together fragments.” Structural features allow us to quantify this, seeing whether the model has learned to prefer copying or generation in particular syntactic environments.

 We incorporate two structural measures into our model: the root distance of word $w_i$, denoted as ${\Droot(w_i)}$ and the edge distance between word $w_{i-1}$ and $w_i$, denoted as ${\Dedge(w_{i-1}, w_i)}$. These measures are calculated on parse trees of generated summaries.\footnote{Parses and part of speech tags are generated by the top-down constituency parser.}
 Root distance is the distance in the parse tree from the current word to the root node, and corresponds to the depth of the word in the parse tree. This measure will tell us if there is an association between depth in the tree and the decision to copy or generate. Edge distance is the number of intervening edges between the current and previous word in the summary. Edge distance will be smaller within a constituent than across two constituents. This measure allows us to test whether the decision to copy or generate is associated with the size of the syntactic boundary between words.

\textbf{Part of Speech:} In addition to structure, the summarization model may condition its decision to copy or generate on the syntactic category of the most recently generated word. For example, in our preliminary qualitative observations of the CNN/DailyMail model, we found that ${\pgen}$ was higher when decoding punctuation, main verbs and conjunctions. To test the association between part-of-speech and ${\pgen}$ formally, we include the part-of-speech label of the current word in our model.

\subsubsection{CNN/DailyMail Results}

We predicted $\pgen$ using four single feature-set linear models, and a single linear model including all features. We conducted ANOVA tests on all combinations of nested models, and found that each set of features significantly improves the $\pgen$ model (all $p < 0.00001$; see Table~\ref{tab:pgen_model}).

\textbf{Entropies:} The coefficients for the model-internal entropy measures $\hgen$ and $\hcopy$ intuitively indicate that as uncertainty in the generation distribution increases, the model is less likely to generate, and as uncertainty in the copy distribution increases, the model is less likely to copy; these relationships were previously explored in Section \ref{section:experiment_token_level}.

The three language model entropy estimates are significantly associated with ${\pgen}$. However, the coefficients are all very small and this feature set individually does the poorest job of explaining ${\pgen}$\!'s variance of all the sets we analyzed. This could be due to the fact that, with the exception of the \ngram\ model, the language model entropy estimates come from different training data than the summarization model. Regardless, while language model entropies significantly improved $\pgen$ prediction, the other feature sets showed a much stronger relationship with ${\pgen}$. Therefore we do not focus on language model entropies in subsequent sections.

\begin{table}[!t]
    \centering
    \small
    \begin{tabular}{ c c c } 
    \hline
     Feature Set & Feature & $\beta$ \\ 
     \hline\hline
    Summ. Model Entropies & ${\hgen}$ &  -0.052 \\ 
    ($R^2$ = 0.274) & ${\hcopy}$ & 0.035 \\ 
    \hline
     LM Entropies & ${\hlstm}$ &  0.009 \\ 
     ($R^2$ = 0.140) & ${\hparser}$ &  0.003 \\ 
     & ${\hngram}$ &  0.009 \\ 
    \hline
     Structural Features & ${\Dedge(w_{i-1}, w_i)}$ & 0.018 \\ 
     ($R^2$ = 0.204) & ${\Droot(w_i)}$ & -0.031 \\
    \hline
      & \$ &  -0.130 \\
      & UH &  -0.118 \\ 
      & \# & -0.116 \\
     Part of Speech & NNP &  -0.111 \\
     ($R^2$ = 0.593) & WRB & 0.156 \\
     & : &  0.254 \\
     & , & 0.269 \\
     & . &  0.636 \\
     \hline\hline
     Full Model $R^2$: 0.648  \\
     \hline
    \end{tabular}
    \caption{Table of slope coefficients ${\beta}$ in the full linear model of ${\pgen}$ in the CNN/DailyMail model. Reported below the name of the feature set is the adjusted ${R^2}$ of a model fit only to that feature set. The eight part of speech tags with the largest magnitude ${\beta}$ are reported. All reported ${\beta}$ are significant via t-test (all ${p < 0.00001}$).}
    \label{tab:pgen_model}
\end{table}

\textbf{Structural Features:} Both structural features are significantly associated with ${\pgen}$. A model fit using only $\Dedge$ and $\Droot$ explains 20\% of ${\pgen}$'s variance ($R^2 = 0.204$). Edge-distance is positively associated with ${\pgen}$, meaning the larger the syntactic boundary between the previous and current word, the more likely the summarization model is to generate. This provides evidence that the model has some knowledge of syntactic boundaries, and uses the generation component as a means of joining together clauses, in line with the observations of \citet{see-etal-2017-get}. We also find that distance to the root node of the parse is negatively associated with ${\pgen}$. This means that words which are higher in the parse tree are more likely to be generated than copied. Conversely, this means that generated components are unlikely to be associated with complex, deeply nested phrasing, suggesting \textbf{the generation component only produces simple shallow substitutions} rather than structurally complex paraphrases or even simple substitutions that modify structurally complex copied elements.

\textbf{Part-of-Speech:}
The part of speech tags with the highest negative association with ${\pgen}$ (i.e.\ those most likely to be copied) are \$ (currency symbols), UH (interjection), \# (pound symbol), followed by NNP (singular proper nouns). These results are perhaps unsurprising, as interjections and proper nouns are difficult to paraphrase and are often out-of-vocabulary in the generation component of the summarization model. \$ and \# serve as prefixes to numerical values which cannot be faithfully paraphrased and therefore should be copied directly from the source text. The tag for a cardinal number (CD) also has a relatively strong negative correlation with ${\pgen}$ (${\beta}$ = -0.088).

The part-of-speech tags with the highest positive association with ${\pgen}$ (i.e.\ those most likely to be generated) are “.” (sentence-final punctuation), “,” (comma), “:” (colon), and WRB (wh-adverbs, such as "where" or "when”). All of these categories can link two clauses or complete sentences, consistent with the ``stitching'' hypothesis of \citet{see-etal-2017-get}.

\begin{figure}
    \centering
    \includegraphics[width=\linewidth]{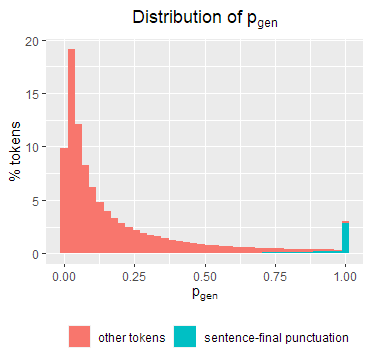}
    \caption{Distribution of ${\pgen}$ across all tokens in the test split of the CNN/DailyMail corpus. Sentence-final punctuation makes up 5\% of tokens in the dataset, which accounts for 22\% of ${\pgen}$'s mass}
    \label{fig:pgen_dist}
\end{figure}
The mean ${\pgen}$ value of all tokens in the test dataset was 0.204, while the mean ${\pgen}$ value for sentence-final tokens was 0.915. Further inspection of the ${\pgen}$ distribution reveals a cluster of outliers at ${\pgen}$ = 1.0. Figure \ref{fig:pgen_dist} shows the distribution of ${\pgen}$ values. We find that, of all tokens with ${\pgen} > 0.95$, 92.1\% are sentence-final punctuation. Despite making up 5\% of all tokens, periods account for 22.1\% of the total mass of ${\pgen}$ in the dataset. This suggests that sentence final punctuation is entirely controlled by the generation distribution. Additionally, we find that of all 5-grams in generated-summaries ending with sentence-final punctuation, 52\% are also present in the article text, compared to 12\% in the reference summaries. Despite the large ${\pgen}$ values exhibited by sentence-final punctuation, the model only generates punctuation in novel contexts less than half of the time, suggesting that \textbf{even when the model heavily utilizes its generative component, it essentially generates a copy of the source text}.

Our explanatory model of ${\pgen}$ shows that model entropy, syntactic depth, syntactic boundary size, and part-of-speech are associated with ${\pgen}$. The strongest predictor of ${\pgen}$ is the part-of-speech of the current word, with copying most strongly associated with numbers, number prefixes and proper nouns, and generation most strongly associated with punctuation. We find that sentence-final punctuation is handled almost entirely by the generative component of the model, despite the fact that sentence-final punctuation occurs in novel contexts less than half of the time.

\subsubsection{XSum Results}
\label{sec:pgen_probe_xsum}

Overall, we find that the variance of $\pgen$ in the XSum model is well explained by model-internal entropy, and relatively poorly explained by linguistic features. We believe this is driven by the categorically different behaviors of each model.%
\footnote{The full table of model coefficients can be found in Table \protect\ref{tab:pgen_model_xsum} of Appendix~C.} 
While the CNN/DailyMail model only uses the generative component to join together copied constituents, the generative component dominates the XSum model's behavior. The mean $\pgen$ value across all tokens in the XSum dataset was 0.828, compared to 0.204 in the CNN/DailyMail dataset. While the structural features ${\Dedge(w_{i-1}, w_i)}$ and ${\Droot(w_i)}$ explained 20.4\% of the variance of $\pgen$ in the CNN/DailyMail model, these features only explain 4.9\% of the variance in the XSum model. Part of speech also does a poorer job of explaining the variance in XSum's $\pgen$. While part of speech explains 59.3\% of the variance of $\pgen$ in the CNN/DailyMail model, part of speech tags only explain 23.0\% in the XSum model.

While the CNN/DailyMail model assigned an abnormally high $\pgen$ value to punctuation, we do not observe this behavior in the XSum model. The CNN/DailyMail model appeared to make use of the “.”, “:” and “,” tokens to join together copied sentences, but none of these tokens are a significant predictor of $\pgen$ in the XSum model. This suggests that the XSum model does not use the generation distribution to connect copied clauses. 

While the XSum model appears not to use the copy and generation distributions in the same way as the CNN/DailyMail model, we still observe some clear and intuitive associations between part of speech tags and $\pgen$. In particular, the XSum model appears to use the copy distribution to handle words which are likely to be out-of-vocabulary for the generation distribution. For example, singular and plural proper nouns, interjections and foreign words (NNP, NNPS, UH, and FW respectively) are associated with low values of $\pgen$ (copying), while all types of verbs are associated with large values of $\pgen$ (generation).

We conclude that the CNN/DailyMail model primarily makes use of lexical and syntactic information such as clause boundaries and punctuation to modulate between copying and generation. By contrast, the XSum model primarily relies on the generation distribution, and backs off to the copy distribution at times of high generation uncertainty or high copy certainty, such as when copying a quote or a proper name. 

\subsection{Modifying $\pgen$} \label{section:modifying_pgen}

\begin{table*}[!t]
    \centering
    \small
    \begin{tabular}{{| m{.95\textwidth} |}} \hline
        \textbf{Article Text}: raheem sterling has admitted he is not ready to sign a new contract at liverpool deal despite being offered a \# 100,000-a-week deal to stay with the merseyside club . the 20-year-old wideman edged closer towards an anfield exit after revealing in an interview with the bbc on wednesday that he would have signed for a lot less a year ago . however , despite being one of liverpool 's star men , sterling has struggled to repeat the impressive form he showed for the reds last season . the england international has managed just six goals this season - one less than stoke frontman jon walters - while his conversion rate and minutes per goal ratio have worsened as the graphic below shows . raheem sterling has managed just six goals this season - one less than stoke forward jon walters -lrb- left -rrb- . \\ \hline
        \textbf{Reference}: raheem sterling has revealed he is not ready to sign a new liverpool deal . the reds wideman has struggled to repeat last season's impressive form . the 20-year-old liverpool star has managed just six goals this  season . read: sterling insists he is not a 'money-grabbing 20-year-old' sterling: what he said about contract talks... and what he meant . click here for the latest liverpool news . \\ \hline
        $\mathbf{\pmin=0}$: raheem sterling has admitted he is not ready to sign a new contract . the england international has managed just six goals this season . sterling has managed just six goals this season - one less than stoke forward jon walters . \\ \hline
        $\mathbf{\pmin=0.25}$: raheem sterling has admitted he is not ready to sign a new contract . the england international has managed just six goals this season . {\color{Cerulean} the england international} has managed just six goals this season . \\ \hline
        $\mathbf{\pmin=0.50}$: raheem sterling has admitted he is not ready to sign a new contract . the england international has managed just six goals this season . {\color{Cerulean} the england international} has managed just six goals this season . \\ \hline
        $\mathbf{\pmin=0.75}$: raheem sterling has admitted he is not ready to sign a new {\color{Cerulean} deal} . the {\color{Cerulean} 20-year-old} has {\color{Cerulean} scored} just six {\color{Cerulean} premier league} goals this season . {\color{Cerulean} the 20-year-old} has {\color{Cerulean} scored} just {\color{WildStrawberry} three} goals this season . \\ \hline
        $\mathbf{\pmin=1}$: {\color{Cerulean} man utd face manchester city in the premier league on saturday . the} {\color{WildStrawberry} striker} has {\color{Cerulean} scored} just {\color{WildStrawberry} four} premier league goals this season . the {\color{WildStrawberry} 19-year-old} has scored just {\color{WildStrawberry} three} goals this season . {\color{Cerulean} click here for all the latest premier league news .} \\ \hline
    \end{tabular}
    \caption{Summaries generated for the same randomly selected article with varying values of $\pmin$. Differences from the base model summary are highlighted in {\color{Cerulean} blue}, while non-faithful text is highlighted in {\color{WildStrawberry} red}.}
    \label{tab:min_pgen}
\end{table*}

\subsubsection{Model}

Taking advantage of the smooth interpolation between the generation and copy distribution, we experiment with forcing the CNN/DailyMail model to be more abstractive. This, we expect, will allow us to differentiate between the abstractive \emph{behavior} we observe in the model summaries and the abstractive \emph{capabilities} that the model may have but which it only uses infrequently in practice. We do so by artificially modifying $\pgen$ during decoding. If $\pmin \in [0,1]$ is a parameter that represents the minimum value of $\pgen$ we allow, we then modify $\pgen$ as follows:
\begin{equation}
    \pgen^* = \pmin + (1 - \pmin) \pgen
\end{equation}
This may be viewed as a linear interpolation from the range $[0,1]$ to $[\pmin,1]$. As $\pmin$ grows, the model is forced to rely more heavily on the generation distribution rather than the copy distribution.%
\footnote{We do not run this experiment on the XSum model because it already usually has a large $\pgen$.}

\subsubsection{Results}

We use the same randomly sampled articles used in Section \ref{section:experiment_token_level}.%
\footnote{We see similar patterns in other randomly-sampled summaries, shared in Appendix B.}
Generated summaries for $\pmin$ values in $[0, 0.25, 0.50, 0.75, 1.0]$ can be found in Table \ref{tab:min_pgen}.

Consistent with previous studies, we find that the model is effective at producing grammatical output. At small values of $\pgen$, the model mostly copies sentences verbatim, but shows the ability to cut a sentence short in a grammatical manner. For example, ``raheem sterling has admitted he is not ready to sign a new contract at liverpool deal...'' is shortened to ``raheem sterling has admitted he is not ready to sign a new contract.''

At greater values of $\pgen$, the model continues sentences in a consistent fashion despite substituting nouns or verbs at the beginning or middle of the sentences. For example, ``sterling has managed just six goals...'' at $\pmin=0$ becomes ``the 20-year-old has scored just six premier league goals'' at $\pmin=.75$. However, we do not observe significant paraphrasing beyond these simple substitutions, and at high values of $\pmin$, where the model is forced to rely heavily on the generation distribution, we begin to observe hallucinations where the model inserts inaccurate information about the player's age and the number of goals scored. When $\pmin=1$, the model generates a completely hallucinated sentence, ``man utd face manchester city in the premier league on saturday'' and a non-informative advertisement ``click here for all the latest premier league news.''

\section{Discussion}

Understanding the limitations preventing abstractive summarization models from paraphrasing effectively is our ultimate aim, but answering that question requires an understanding of current models' abstraction capabilities. In this paper, we analyze the abstractions of which the pointer-generator model  \cite{see-etal-2017-get} is capable.

When trained on CNN/DailyMail, we find that sentence truncation is the most common form of paraphrasing. Punctuation tokens are associated with high generation rates and low entropy in the generation distribution. Additionally, high $\pgen$ often results in generating the token that comes next in a phrase already being copied verbatim, suggesting that high $\pgen$ merely gives the model the \emph{option} to generate novel text, but that the model rarely makes use of it. Artificially increasing $\pgen$ does not significantly change this behavior, introducing increased rates of synonym substitution as well as increased rates of non-faithful hallucination.

When trained on XSum, the model makes much less use of the copy mechanism, largely generating novel text with a few exceptions, including the copying of proper nouns and parts of quotations. The model generally produces topical summaries, but ones that aren't necessarily grammatical or faithful to the original article. For example, the randomly selected summary used in Figure \ref{fig:corr_contrib_0} repeats itself and wanders, ``... on the smiler rollercoaster on the smiler rollercoaster in the south west 200 years ago as `phenomenal'''. This comes after a hallucination, ``firefighters are continuing to search for a man'' even though the article describes the rescue from the rollercoaster crash in the past tense. We hypothesize that the phrase ``firefighters are continuing to search'' is a relatively common phrase in news articles that the model learned from the training data. Such frequency biases likely contribute to the faithfulness issues in abstractive summarizers reported in previous literature.

Our results give context to previous observations that summarization model unfaithfulness increases with abstraction \cite{maynez-etal-2020-faithfulness, durmus2020feqa, kryscinski-etal-2020-evaluating} and that abstractive models are prone to output repetition \cite{see-etal-2019-massively, holtzman2020curious}. To faithfully paraphrase, a model must understand both the syntax and the semantics of the original text. The models we studied were able to recognize syntactic boundaries, proper nouns, and noun phrases that could be substituted with synonyms. However, the models didn't appear to comprehend the meaning of the text well enough to generate faithful complex paraphrases. This is unacceptable in high-risk domains such as healthcare; \citet{zhang-etal-2018-learning-summarize} train a model to summarize radiology findings, but only 67\% of their summaries are judged at least as good as human summaries, in a domain where errors can have a major impact on human lives.

In our work, the explicit switch between abstractive and extractive modes enabled us to directly observe the conditions under which abstractive summarization was chosen as a strategy, and to force an abstractive summarization strategy to disentangle paraphrasing behavior from capabilities. We found that the \citet{see-etal-2017-get} model trained on CNN/DailyMail did learn simple forms of paraphrasing, despite the extractive bias of the dataset. We conclude that pointer-generator models are \emph{capable} of simple paraphrasing regardless of training data, even though they \emph{behave} in ways that rely on the frequency biases of the training dataset. However, they also appear \emph{incapable} of producing significant paraphrases that are grammatical, non-repetitive, and faithful to the source document. 
This suggests that using an abstractive-biased dataset alone is not enough for a model to learn robust and faithful paraphrasing strategies. Rather, when trained on XSum, the pointer-generator model seems to simply learn that it should not copy from the source text.
Future work should investigate how either datasets or models can improve the training signal that allows the model to understand the underlying semantics of the source document.

Related to our work, \citet{xu-etal-2020-understanding} 
studied the summarization strategies of state-of-the-art transformer summarization models. Since their models did not contain an explicit copy/generation switch, they used \ngram\ overlap between source documents and summaries as a proxy to measure a summary's ``extractiveness.'' They found a similar result to ours, that high \ngram\ overlap (``copying'') corresponded to low entropy in the decoder's output distribution when the model was trained on CNN/DailyMail.%
\footnote{Their findings are more difficult to interpret when trained on XSum, partially due to the lack of an explicit extractive/abstractive summarization switch in their models.}
Their findings suggest that our results likely generalize to a much broader class of summarization models than the pointer-generator models studied here. 

Finally, \citet{liu2010exploring} found that ROUGE metrics poorly correlate with human evaluations, leading to recent models being evaluated with human judgements, but these evaluations often disagree on what they are measuring, whether it is faithfulness, informativity, or the unqualified ``quality'' of a summary \cite{zhang2018abstractiveness, zhang2020pegasus, dou2020gsum}. Developing best practices on how abstractive summarizers should be evaluated for their paraphrasing ability is another problem we leave for future work.

\section{Conclusion}

In this paper, we presented three experiments that evaluate the abstraction capabilities of the pointer-generator neural summarization model. Our results conclude that on extractive training data, the model uses only simple paraphrasing strategies that truncate sentences at syntactic boundaries, allowing the model to stay grammatically accurate as well as faithful to the source document. We explore two ways to make the model use abstractive summarization strategies: modifying the model so that it relies more heavily on its abstractive component, and training a new model on an abstractive-biased dataset. In both cases, the model shows simple paraphrasing capabilities but frequently generates unfaithful paraphrases. These results highlight current limitations of abstractive summarization, where in lieu of semantic understanding, models must rely on extractive heuristics in order to stay faithful.

\section*{Acknowledgements}

We thank our reviewers for their helpful suggestions. We also thank Esin Durmus, Ryan Benmalek, and Claire Cardie for helpful discussions about abstractive summarization. Finally, we thank Shashi Narayan and Shay Cohen for helping us reproduce their pointer-generator model trained on XSum.

\bibliographystyle{acl_natbib}
\bibliography{bibliography}

\newpage

\onecolumn

\section*{Appendix A. Additional Correlation Contribution Examples}

This appendix includes additional examples of $\CC(\pgen, \hgen)$, the per-token correlation contributions for randomly selected summaries.

\begin{figure*}[h!]
    \centering
    \begin{subfigure}[b]{\textwidth}
        \centering
        \includegraphics[width=\textwidth]{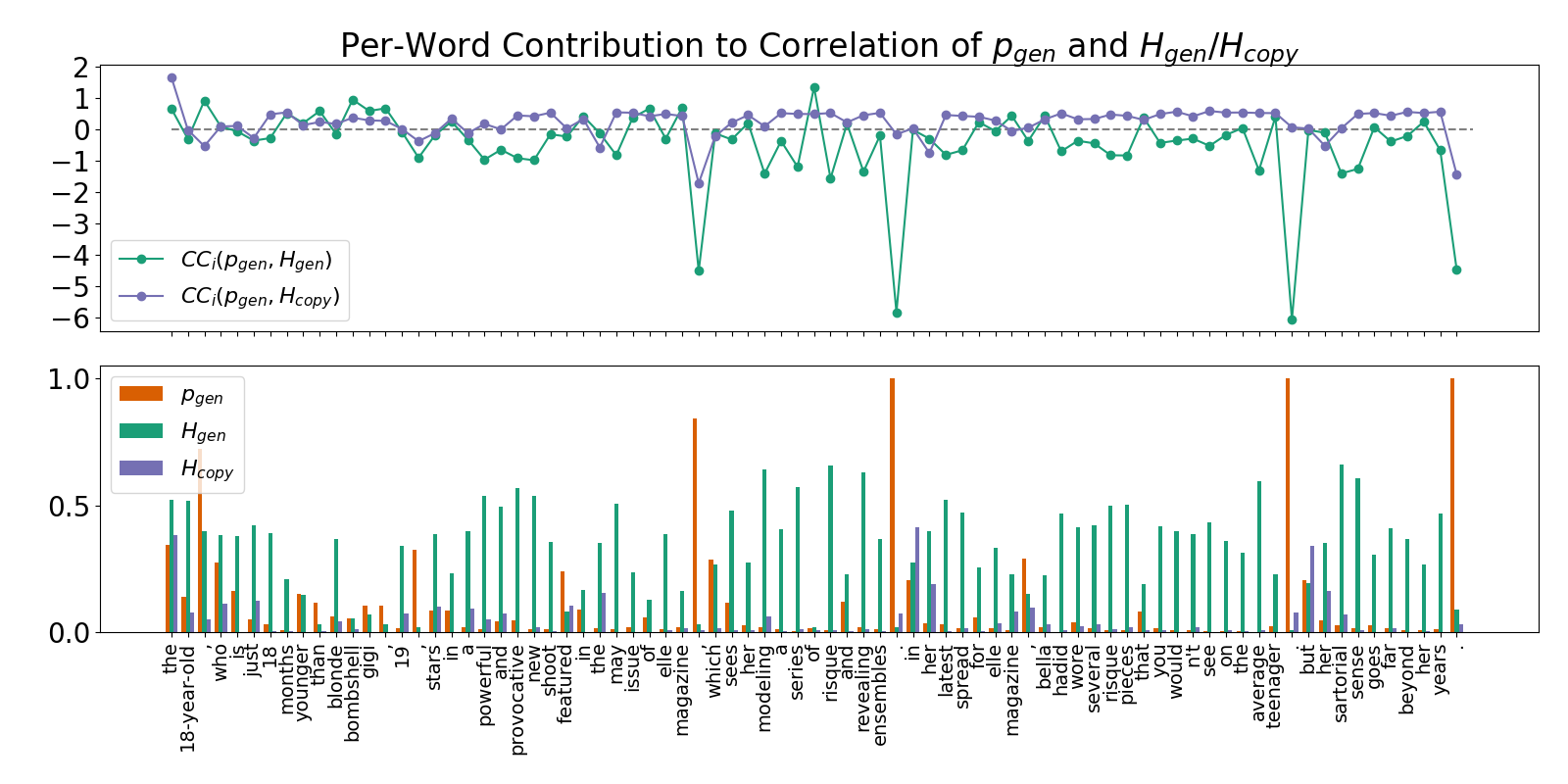}
        \caption{CNN/DailyMail Example 2}
    \end{subfigure}
    
    \begin{subfigure}[b]{\textwidth}
        \centering
        \includegraphics[width=\textwidth]{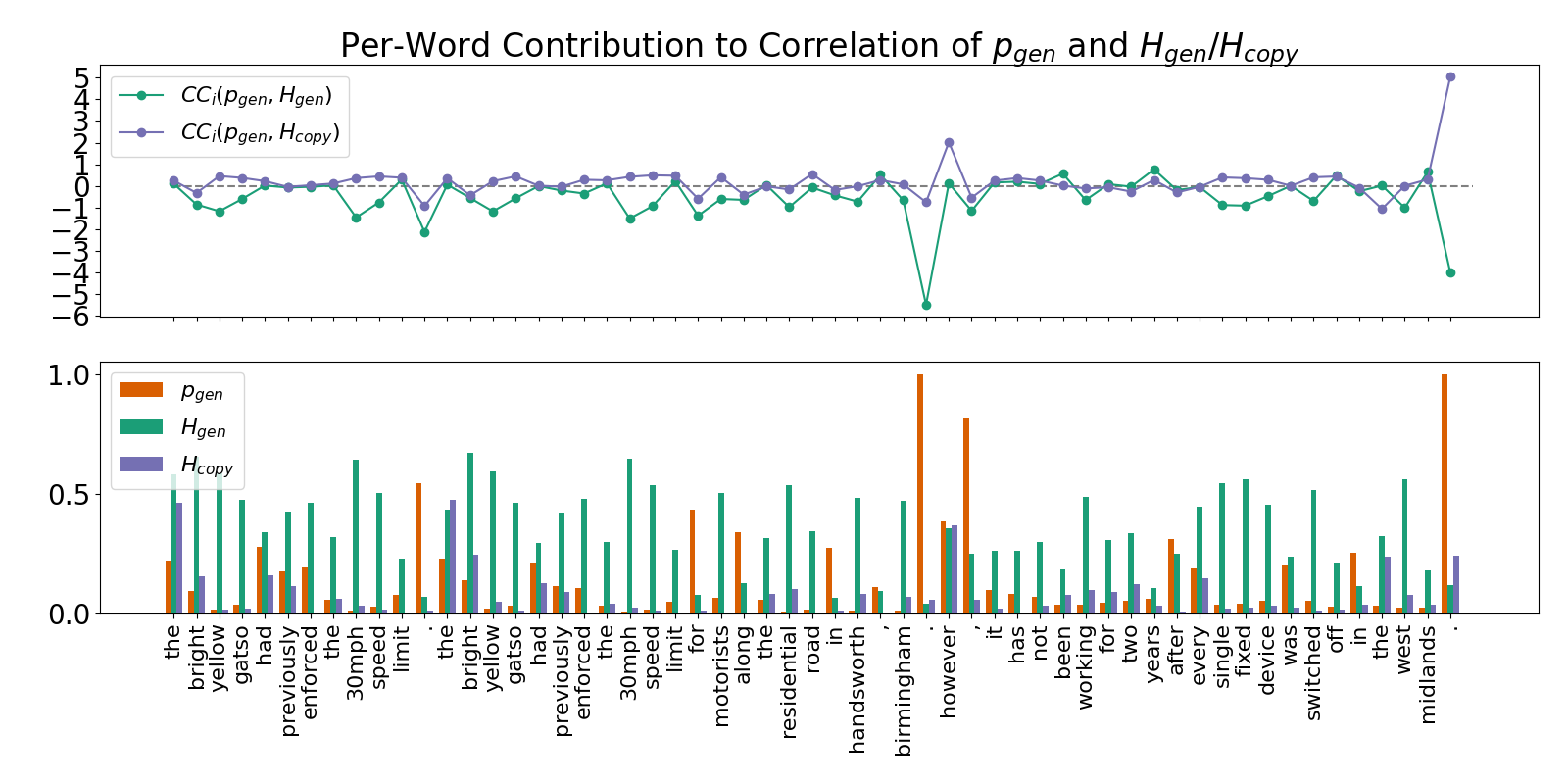}
        \caption{CNN/DailyMail Example 3}
    \end{subfigure}
    \caption{Bar plot of per-token $\pgen$ and entropy of the generation distribution (purple) and copy distribution (blue), plotted under correlation contributions $\CC(\pgen, \hgen)$ (purple) and $\CC(\pgen, \hcopy)$ (blue) for a randomly-sampled CNN/DailyMail test summaries.}
\end{figure*}
    
\begin{figure*}
    \centering
    \begin{subfigure}[b]{\textwidth}
        \centering
        \includegraphics[width=\textwidth]{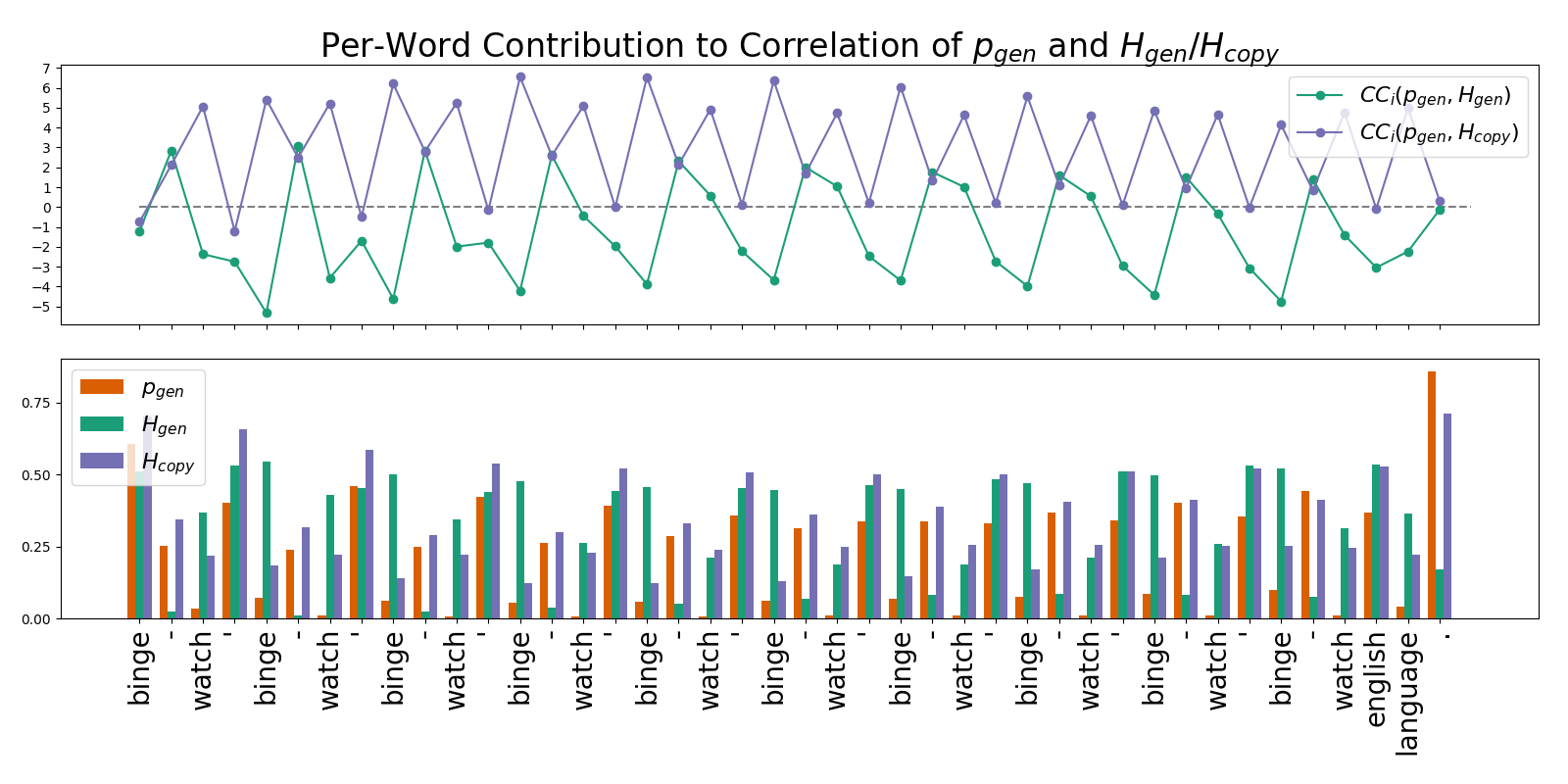}
        \caption{XSum Example 2}
    \end{subfigure}
    
    \begin{subfigure}[b]{\textwidth}
        \centering
        \includegraphics[width=\textwidth]{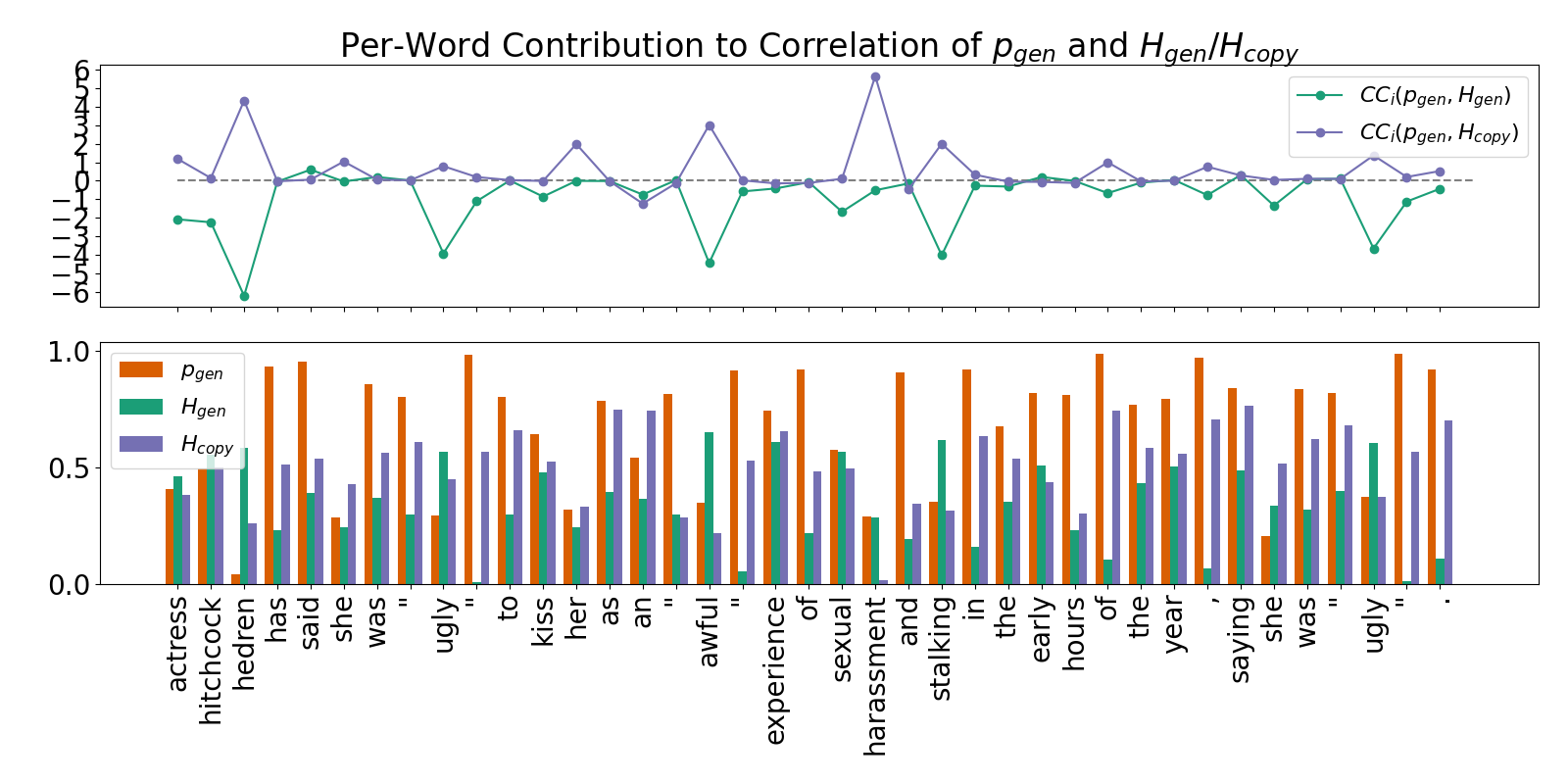}
        \caption{XSum Example 3}
    \end{subfigure}
    
    \caption{Bar plot of per-token $\pgen$ and entropy of the generation distribution (purple) and copy distribution (blue), plotted under correlation contributions $\CC(\pgen, \hgen)$ (purple) and $\CC(\pgen, \hcopy)$ (blue) for a randomly-sampled XSum test summaries.}
    \label{fig:corr_contrib_1}
\end{figure*}

\clearpage

\section*{Appendix B. Additional Summaries With Modified $\pmin$}

This appendix contains additional examples of randomly selected summaries generated with modified values of $\pmin$.

\begin{table*}[h!]
    \centering
    \small
    \begin{tabular}{{| m{.95\textwidth}|}} \hline
        \textbf{Article Text}: gigi hadid 's younger sister bella hadid is quickly catching up with her star sibling 's phenomenal fashion success , carving out an impressive career for herself within the industry . the 18-year-old , who is just 18 months younger than blonde bombshell gigi , 19 , stars in a powerful and provocative new shoot featured in the may issue of elle magazine , which sees her modeling a series of risque and revealing ensembles . in one image , the dark-haired beauty poses in a sheer fleur du mal lace bodysuit , with her arms thrown casually above her head . scroll down for video . it 's good to be a model : bella hadid models a tantalizing \$ 895 fleur du mal lace bodysuit and \$ 1,250 giuseppe zanotti boots . giuseppe zanotti cowboy boots . on sale for \$ 500 at 6pm ! visit site . she may be just 18 , but her sartorial sense goes far beyond her years . and in her latest spread for elle magazine , bella hadid wore several risque pieces that you would n't see on the average teenager . but then again , the budding supermodel is n't a normal teen [...] \\ \hline
        \textbf{Reference}: bella, 18, is the younger sister of guess campaign star gigi hadid, 19 . the rising star poses in a series of provocative outfits for the may issue of elle . fellow fashion favorite hailey baldwin also features in the issue, appearing in her own separate shoot and interview . \\ \hline
        $\mathbf{\pmin=0}$: the 18-year-old , who is just 18 months younger than blonde bombshell gigi , 19 , stars in a powerful and provocative new shoot featured in the may issue of elle magazine , which sees her modeling a series of risque and revealing ensembles . in her latest spread for elle magazine , bella hadid wore several risque pieces that you would n't see on the average teenager . but her sartorial sense goes far beyond her years . \\ \hline
        $\mathbf{\pmin=0.25}$: the 18-year-old , who is just 18 months younger than blonde bombshell gigi , 19 , stars in a powerful and provocative new shoot featured in the may issue of elle magazine , which sees her modeling a series of risque and revealing ensembles . in her latest spread for elle magazine , bella hadid wore several risque pieces that you would n't see on the average teenager . \\ \hline
        $\mathbf{\pmin=0.50}$: the 18-year-old is just 18 months younger than blonde bombshell . {\color{Cerulean} she is just 18 months younger than blonde bombshell gigi , 19 . she is just 18 months younger than blonde bombshell gigi .} \\ \hline
        $\mathbf{\pmin=0.75}$: the 18-year-old is just 18 months younger than blonde bombshell .{\color{Cerulean} she is just 18 months younger than blonde bombshell . she is just 18 months younger than blonde bombshell . she is just 18 months younger than blonde bombshell .} \\ \hline
        $\mathbf{\pmin=1}$: the {\color{Cerulean} [UNK]} , {\color{WildStrawberry} 33} , {\color{WildStrawberry} has been dating for three years} . she is the daughter of [UNK] and [UNK] . she is the daughter of [UNK] 's daughter . her {\color{WildStrawberry} new album is priced at} \$ [UNK] -lrb- \# [UNK] -rrb- . \\ \hline
    \end{tabular}
    \label{tab:min_pgen_2}
    
    \bigskip
    
    \begin{tabular}{{| m{.95\textwidth}|}} \hline
        \textbf{Article Text}: the owners of this house better not leave too quickly , after a speed camera was angled directly on their front door . the bright yellow gatso had previously enforced the 30mph speed limit for motorists along the residential road in handsworth , birmingham . however , it has not been working for two years after every single fixed device was switched off in the west midlands . big brother is watching : a speed camera has been turned round and is pointing at this house in birmingham , west midlands . the speed camera has not been working for more than two years [...] \\ \hline
        \textbf{Reference}: speed camera discovered pointing at house in handsworth, birmingham . fixed cameras  switched off across the west midlands in spring of 2013 . site is not going to be part of a new trial using digital technology . obsolete camera may now be taken down after engineers examine device . \\ \hline
        $\mathbf{\pmin=0}$: the bright yellow gatso had previously enforced the 30mph speed limit . the bright yellow gatso had previously enforced the 30mph speed limit for motorists along the residential road in handsworth , birmingham . however , it has not been working for two years after every single fixed device was switched off in the west midlands . \\ \hline
        $\mathbf{\pmin=0.25}$: the bright yellow gatso had previously enforced the 30mph speed limit for motorists along the residential road in handsworth , birmingham . it has not been working for more than two years after every single fixed device was switched off in the west midlands . {\color{Cerulean} the speed camera} has not been working for more than two years . \\ \hline
        $\mathbf{\pmin=0.50}$: the bright yellow gatso had previously enforced the 30mph speed limit . it has not been working for more than two years . it has not been working for more than two years after every single fixed device \textbf{.} \\ \hline
        $\mathbf{\pmin=0.75}$: the bright yellow gatso had previously enforced the 30mph speed limit . it has not been working for more than two years . it has not been working for more than two years . it has not been working for more than two years . \\ \hline
        $\mathbf{\pmin=1}$: {\color{WildStrawberry} warning : graphic content} . {\color{Cerulean} it is believed to have been in the past of} the past two years . {\color{Cerulean} it is believed to have been in the past of} the past two years . \\ \hline
    \end{tabular}
    
    \caption{Summaries generated for additional randomly selected articles from CNN/DailyMail with varying values of $\pmin$. Differences from the base model summary are highlighted in {\color{Cerulean} blue}, while non-faithful text is highlighted in {\color{WildStrawberry} red}.}
\end{table*}

\begin{table*}[h!]
    \centering
    \small
    \begin{tabular}{{| m{.95\textwidth} |}} \hline
        \textbf{Article Text}: meaning " to watch a large number of television programmes ( especially all the shows from one series ) in succession " , it reflects a marked change in viewing habits , due to subscription services like netflix . lexicographers noticed that its usage was up 200 \% on 2014 . other entries include dadbod , ghosting and clean eating . helen newstead , head of language content at collins , said : " the rise in usage of ' binge - watch ' is clearly linked to the biggest sea change in our viewing habits since the advent of the video recorder nearly 40 years ago . " it 's not uncommon for viewers to binge - watch a whole season of programmes such as house of cards or breaking bad in just a couple of evenings - something that , in the past , would have taken months - then discuss their binge - watching on social media . " those partaking in binge - watching run the risk of dadbod , one of ten in the word of the year list [...]
        the list of collins ' words of the year offers a fascinating snapshot of the ever - changing english language , " said newstead . those words that remain popular could be included in the next print edition of the collins english dictionary , due in 2018 . \\ \hline
        \textbf{Reference}: collins english dictionary has chosen binge-watch as its 2015 word of the year. \\ \hline
        \textbf{Summary}: binge - watch ' binge - watch ' binge - watch ' binge - watch ' binge - watch ' binge - watch ' binge - watch ' binge - watch ' binge - watch ' binge - watch english language . \\ \hline
        \textbf{Summary with Coverage}: the risk of binge - watch ' binge - watch english language is {\color{WildStrawberry} " clearly uncommon "} , {\color{WildStrawberry} according to a list of entries} from the {\color{WildStrawberry} collins english media recorder of the year list} . \\ \hline
    \end{tabular}
    \label{tab:min_pgen_xsum}
    
    \bigskip
    
    \begin{tabular}{{| m{.95\textwidth} |}} \hline
        \textbf{Article Text}: writing in her autobiography , she claimed the director " threw himself " on top of her in the back of his limousine and tried to kiss her . the actress described the encounter as " an awful , awful moment " . hedren added that she did n't tell anyone because " sexual harassment and stalking were terms that did n't exist " in the early 1960s . she continued : " besides , he was alfred hitchcock [...] 
        the actress , now 86 , made the claims in her autobiography tippi : a memoir , which is published in november . she has spoken in the past about the director 's alleged treatment of her , but has gone into more detail in the memoir . hedren described a later encounter in hitchcock 's office where the director " suddenly grabbed " her and " put his hands " on her . she wrote : " it was sexual , it was perverse , and it was ugly , and i could n't have been more shocked and more repulsed . " [...]
        the actress said hitchcock then made her life difficult , refusing to submit her work for the oscar nominations or let her take on other acting roles while he still had her under contract [...]
        \\ \hline
        \textbf{Reference}: actress tippi hedren has claimed alfred hitchcock sexually harassed her while they worked together in the 1960s. \\ \hline
        \textbf{Summary}: {\color{WildStrawberry} actress hitchcock hedren} has said she was " ugly " to kiss her as an " awful " experience of sexual harassment and stalking in the early hours of the year , saying she was " ugly " . \\ \hline
        \textbf{Summary with Coverage}: {\color{WildStrawberry} actress hitchcock hitchcock} , {\color{WildStrawberry} best known by the director} of the oscar - winning director , {\color{WildStrawberry} has died} at the age of 86 , the actress has announced on her return to the memoir . \\ \hline
    \end{tabular}
    
    \caption{Summaries generated for additional randomly selected articles from XSum with varying values of $\pmin$. Summaries with coverage enabled also included. Non-faithful text is highlighted in {\color{WildStrawberry} red}}
\end{table*}

\newpage

\section*{Appendix C. Explanatory $\pgen$ Model for XSum Dataset}
\begin{table*}[h!]
    \centering
    \small
    \begin{tabular}{ c c c } 
    \hline
     Feature Set & Feature & $\beta$ \\ 
     \hline\hline
    Summ. Model Entropies & ${\hgen}$ &  -0.099 \\ 
    ($R^2$ = 0.476) & ${\hcopy}$ & 0.093 \\ 
    \hline
     LM Entropies & ${\hlstm}$ &  0.009 \\ 
     ($R^2$ = 0.123) & ${\hparser}$ &  0.003 \\ 
     & ${\hngram}$ &  -0.013 \\ 
    \hline
     Structural Features & ${\Dedge(w_{i-1}, w_i)}$ & -0.005 \\ 
     ($R^2$ = 0.049) & ${\Droot(w_i)}$ & -0.001 \\
    \hline
      & NNPS &  -0.166 \\
      & FW &  -0.162 \\ 
      & UH & -0.143 \\
     Part of Speech & NNP &  -0.089 \\
     ($R^2$ = 0.230) & VBD & 0.174 \\
     & LS & 0.179 \\
     & VBN & 0.178 \\
     & WP\$ & 0.193 \\
     \hline\hline
     Full Model $R^2$: 0.547  \\
     \hline
    \end{tabular}
    \caption{Table of slope coefficients ${\beta}$ in the full linear model of ${\pgen}$ in the XSum model. Reported below the name of the feature set is the adjusted ${R^2}$ of a model fit only to that feature set. The eight part of speech tags with the largest magnitude ${\beta}$ are reported. All reported ${\beta}$ are significant via t-test (all ${p < 0.00001}$).}
    \label{tab:pgen_model_xsum}
\end{table*}

\end{document}